\documentclass{article} 

\usepackage{iclr2027_conference}
\usepackage{times}
\iclrfinalcopy

\usepackage{amsmath,amsfonts,bm}

\def\eqref#1{equation~\ref{#1}}

\def\1{\bm{1}}

\DeclareMathAlphabet{\mathsfit}{\encodingdefault}{\sfdefault}{m}{sl}
\SetMathAlphabet{\mathsfit}{bold}{\encodingdefault}{\sfdefault}{bx}{n}

\usepackage{hyperref}
\usepackage{url}
\usepackage[flushleft]{threeparttable}
\usepackage{booktabs}
\usepackage{makecell}
\usepackage{multirow}
\usepackage[most]{tcolorbox}
\usepackage{fvextra}
\usepackage{xcolor}
\usepackage{cleveref}
\usepackage{wrapfig}
\usepackage{booktabs}
\usepackage{graphicx}
\usepackage{threeparttable}
\usepackage{needspace}
\usepackage{placeins}
\usepackage{subcaption}
\usepackage{float}
\usepackage{xcolor}
\definecolor{darkgreen}{RGB}{0,100,0}

\newcommand{\ABBR}{\text{DuoMind}}
\newcommand{\BENCH}{\text{RoboPoly}}

\title{\ABBR{}: Enabling Distributed Multi-Robot Coordination with Semantic Communication}

\author{
\textbf{Hanchu Zhou}$^{1}$\thanks{\texttt{hczhou@ucdavis.edu}}
\hspace{0.8em}
\textbf{Dechen Gao}$^{1}$
\hspace{0.8em}
\textbf{Hang Wang}$^{2}$
\hspace{0.8em}
\textbf{Brendan Lynch}$^{3}$ 
\hspace{0.8em}
\textbf{Boqi Zhao}$^{1}$ \\[-0.1em]
\hspace{0.06em}
\textbf{Qiyao Ma}$^{1}$
\hspace{0.8em}
\textbf{Raman Goyal}$^{3}$
\hspace{0.8em}
\textbf{Junshan Zhang}$^{1}$\thanks{Corresponding author: \texttt{jazh@ucdavis.edu}}
\\[0.45em]
$^{1}$University of California, Davis
\hspace{1.2em}
$^{2}$Microsoft Research
\hspace{1.2em}
$^{3}$Analog Devices
}

\begin{document}

\maketitle
\lhead{}

\begin{abstract}

Vision–language models (VLMs) and vision–language–action models (VLAs) have recently driven rapid progress in general-purpose robots, yet most progress has focused on single-robot settings. Extending these capabilities to multi-robot systems remains challenging because robots must coordinate long-horizon behaviors while maintaining reliable, fine-grained execution. We introduce \ABBR{}, a distributed hierarchical framework for multi-robot coordination through semantic communication. Each robot uses a VLA-based action model for low-level execution and a VLM-based orchestrator for high-level reasoning and inter-agent coordination. At each planning step, the orchestrator at each robot reasons over the task instruction, local observations, and messages received from other robots. It then generates low-level instructions for the action model and semantic messages for peer robots. This architecture exploits the complementary strengths of pretrained models by combining the semantic reasoning capabilities of VLMs with the precise action-generation capabilities of VLAs. To address the scarcity of benchmarks for multi-robot coordination, we further develop \BENCH{}, a benchmark comprising  long-horizon manipulation tasks that require coordinated, closed-loop execution under distributed control. Experiments on \BENCH{} and RoboTwin demonstrate that \ABBR{} improves multi-robot task performance, while ablation studies confirm the contributions of hierarchical orchestration and semantic communication. More details are available on our \href{https://hanchuzhou.github.io/duomind_project_page/}{\textcolor{darkgreen}{project page}}.

\end{abstract}

\begin{figure*}[h]
    \centering
    \includegraphics[width=1.0\linewidth]{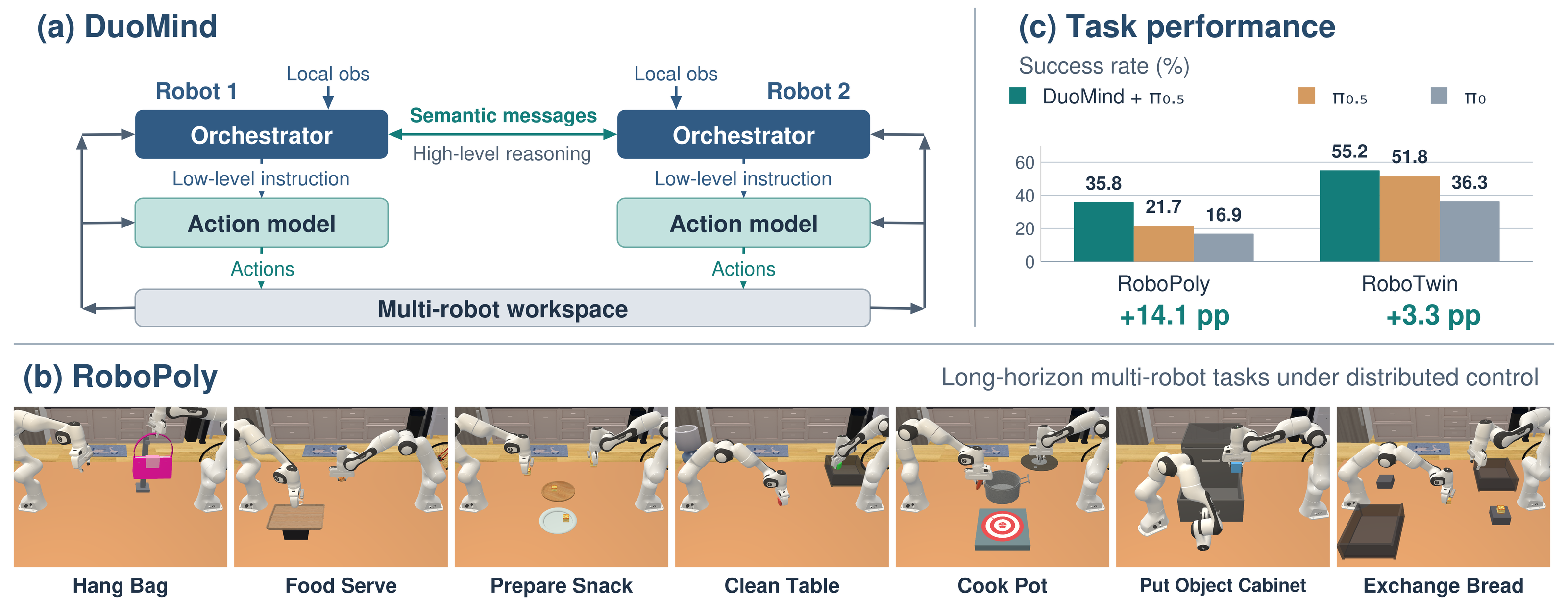}
    \vspace{-0.25in}
    \caption{(a) An illustration of \ABBR{}, a distributed framework combining an orchestrator and action model for multi-robot coordination. (b) \BENCH{} is a benchmark we develop for long-horizon coordination under distributed robot control.
    (c) \ABBR{} consistently outperforms the  baselines across both benchmarks.
}
    \label{fig:illustration} 
    \vspace{-0.1in}
\end{figure*}

\section{Introduction}
\label{sec:intro}

Recent advances in vision-language models (VLMs) and vision-language-action models (VLAs) have substantially expanded the capabilities of general-purpose robots. VLMs \citep{beyer2024paligemma,yang2025qwen3} provide strong visual understanding and high-level reasoning, while VLAs translate visual observations and natural-language instructions into robot actions \citep{zitkovich2023rt,kim2024openvla,black2024pi_0,black2025pi05, team2025gemini}. Building on these complementary capabilities, recent design of robot agents \citep{song2025hume,black2025pi05,team2025gemini,intelligence2025pi} combines a high-level embodied reasoning model with a low-level VLA policy. In this design, the reasoning model aims to understand the environment and then decompose complex tasks into more specific instructions, while the VLA executes these instructions through fine-grained robot actions. Such a separation of functionalities allows the high-level model to focus on semantic understanding and long-horizon planning, while the VLA specializes in executing detailed instructions with precise low-level control.

Notably, most of the above progress has focused on a single robot acting in isolation, but many practical tasks are infeasible for a single agent and inherently require multi-robot collaboration, and others can benefit from multiple robots working together to improve efficiency. Effective multi-robot coordination requires each robot to construct appropriate subgoals and continuously adapt its behavior as the shared workspace evolves. For many practical settings, distributed execution \citep{amirkhani2022consensus,lo2024learning,chen2024decentralized} is particularly desirable, where each robot makes decisions primarily from its own observations rather than relying on a centralized controller, making the system more scalable as the number of robots increases \citep{nayak2023scalable}. Yet combining distributed multi-agent decision making with complex robotic manipulation creates a particularly challenging coordination problem, requiring robots to reason about both team-level cooperation and precise low-level task execution.

Conventional multi-agent learning and planning methods are often studied in relatively simple environments \citep{feng2026multi,li2024language,zou2025latent,wong2023deep,liang2025llm}, with limited physical interaction and shorter task horizons. More recently, language-model-based robotic systems have leveraged the high-level reasoning capabilities of LLMs to decompose complex objectives and schedule robot behaviors \citep{chen2024scalable,he2026closed,song2025leveraging,liu2025coherent,mandi2024roco}. However, most of these systems focus primarily on high-level planning, while low-level control is executed through simply predefined action primitives or rule-based skills. An important gap  remains between high-level multi-agent reasoning and end-to-end robot control. In addition, existing benchmarks provide limited support for systematic evaluation of multi-robot coordination, leaving the field without a comprehensive testbed for measuring such capabilities.

To address  the above challenges, we introduce \ABBR{}, a distributed hierarchical framework for multi-agent robotic coordination, together with \BENCH{}, a benchmark designed specifically for long-horizon multi-robot coordination. Each robot in \ABBR{} operates with a VLM as its orchestrator for high-level reasoning and a VLA as its action model for low-level control. At each planning step, each orchestrator reasons over the high-level instruction, its local observation, and the shared message from other robots. It then generates both a low-level instruction for the action model and a structured message for other robots. These two outputs serve complementary roles. The low-level instruction guides the action model to produce fine-grained robot actions, while the message communicates the robot's intention, subgoal, and task belief, allowing each agent to remain aware of its peers despite partial observability \citep{liang2025review} in the distributed setting. In this way, \ABBR{} provides an agentic harness around the action model: the orchestrator dynamically reasons and invokes the action model to realize long-horizon coordinated behavior at a relatively low frequency, while the action model focuses on high-frequency execution. Importantly, this decoupled design allows \ABBR{} to capitalize on the strong pretrained capabilities of existing VLM and VLA backbones. Rather than heavily fine-tuning models specifically for multi-robot coordination, high-level coordination and low-level manipulation can be largely handled by off-the-shelf VLM models, requiring only limited adaptation to the multi-robot setting. To systematically evaluate the coordination capabilities, we introduce \BENCH{}, which provides challenging long-horizon manipulation tasks in which multiple robots must coordinate and complete the task through closed-loop physical execution rather than high-level planning alone. We further evaluate \ABBR{} across diverse cooperative scenarios and systematically study the contributions of hierarchical design and inter-agent communication.

The main contributions of this paper are summarized as follows:
\begin{itemize}
\item We introduce \ABBR{}, a distributed hierarchical framework for end-to-end multi-agent robotic coordination. \ABBR{} utilizes a VLM as the orchestrator at each robot for high-level reasoning and a VLA as the action model for low-level control, enabling inter-agent semantic communication with natural language. This design harnesses the strong reasoning and control capabilities of existing foundation models to enable effective coordination in complex multi-robot tasks.

\item We develop \BENCH{}, a benchmark for long-horizon multi-robot coordination under distributed observations and control. The benchmark emphasizes closed-loop physical execution and includes tasks that are challenging for a single agent, requiring multiple agents to coordinate and schedule their behaviors throughout task execution.

\item We conduct extensive experiments and ablation studies to evaluate \ABBR{} across challenging cooperative manipulation tasks on both \BENCH{} and RoboTwin. Our results demonstrate the effectiveness of the proposed hierarchical design in multi-robot coordination and further quantify the contributions of integrating orchestrator and inter-agent communication.

\end{itemize}

\section{\ABBR{}: Distributed Hierarchical Multi-Agent Framework}
\label{sec:framework}

We introduce \ABBR{}, a distributed hierarchical framework for multi-agent robotic coordination. We first present the VLM-based orchestrator for high-level planning (\ref{sec:embodied reasoning}). We then describe how we bridges high-level planning and low-level control via embodied reasoning and agentic harness (\ref{sec:bridge}). Finally, we highlight the structured inter-agent communication (\ref{sec:information sharing}).

\begin{figure*}[h]
    \centering
    \includegraphics[width=1.0\linewidth]{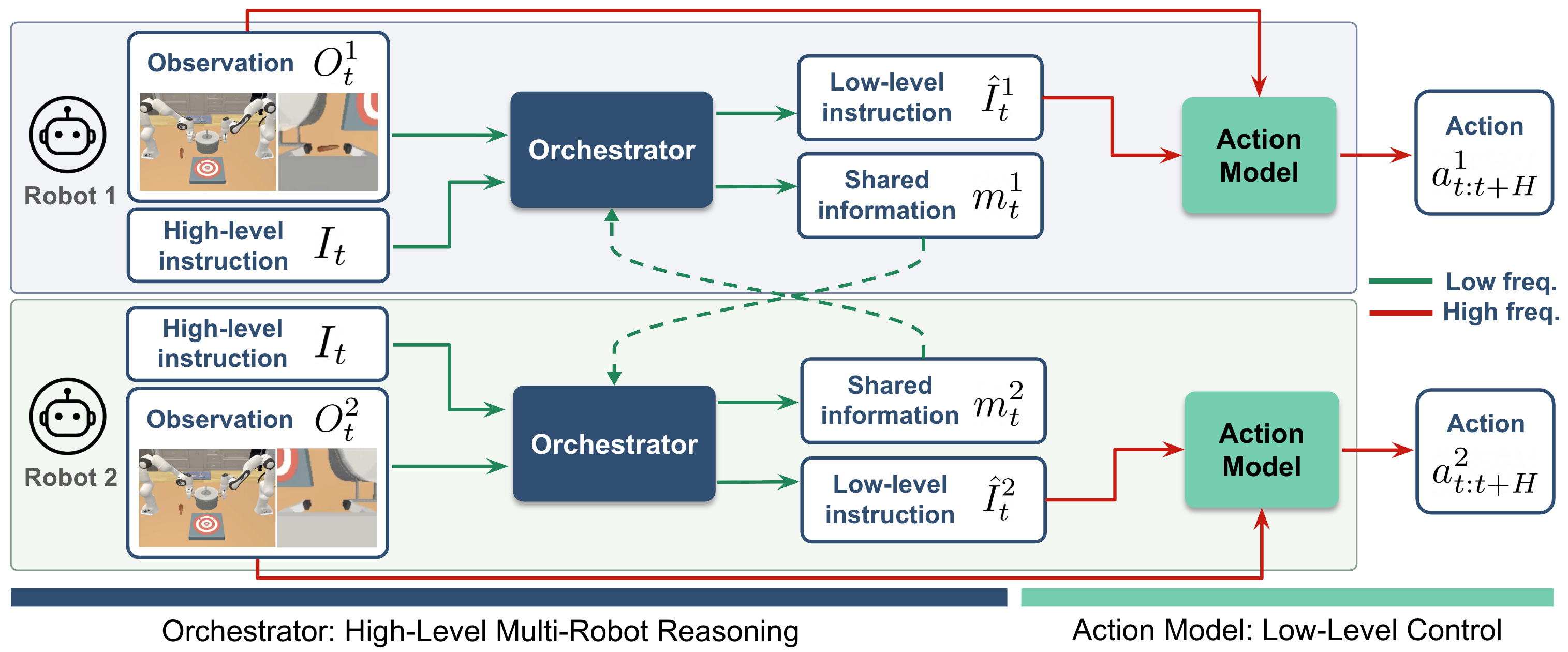}
    \vspace{-0.25in}
    \caption{In \ABBR{}, each robot operates with a VLM as its orchestrator for high-level reasoning and a VLA as its action model for low-level control. At each high-level planning step, the orchestrator at each robot reasons and generates both a low-level instruction for the action model and a structured message for other robots. The action model subsequently follows the low-level instruction and produces fine-grained actions.
}
    \label{fig:illustration} \vspace{-0.1in}
\end{figure*}

\subsection{Multi-Agent Coordination via Embodied Reasoning}
\label{sec:embodied reasoning}

Existing studies on hierarchical robotic architectures have shown that action models are effective at low-level control and instruction following when provided with specific commands \citep{team2025gemini}. To extend this capability toward more general and long-horizon tasks, a promising approach is to introduce a reasoning model as an orchestrator, leveraging its strong visual understanding, semantic grounding, and planning capabilities to generate appropriate instructions for the action model. In distributed multi-agent settings, where task coordination are substantially more complex, we extend this hierarchical paradigm to distributed multi-robot coordination in \ABBR{}.

Specifically, \ABBR{} decomposes multi-agent coordination into two complementary levels: high-level multi-agent reasoning and low-level execution. Each robot employs a VLM as an orchestrator that reasons over the task-level instruction, its local visual observation, and message shared by other agents. The orchestrator determines the robot’s current subtask and translates it into a single-agent, short-horizon instruction for low-level control. The VLA, serving as the action model, then follows the low-level instruction using local observations to generate fine-grained robot actions. This division of responsibility allows each module to focus on the capability for which it is best suited: the orchestrator handles high-level understanding and inter-agent coordination in multi-agent setting, while the action model performs reliable low-level control.

Importantly, this decomposition avoids requiring the action model to directly interpret complex multi-agent state that includes the joint behavior of multiple robots and environment, which can be extremely challenging since action models are primarily trained with single-agent demonstrations. It therefore decouples multi-agent reasoning from low-level control, reducing the need for the action model itself to possess strong reasoning capabilities for multi-agent coordination. Moreover, \ABBR{} does not require the orchestrator, the action model, or the interface between them to be extensively redesigned for multi-agent settings. These foundation models developed for single-agent systems can be reused largely off the shelf, with only lightweight adaptation when necessary. This greatly extends the possibility of deploying \ABBR{} to diverse multi-robot tasks and offers strong compatibility to future upgrades on components.

The design of \ABBR{} naturally supports coordination among multiple robots. However, the complexity of task design, data collection, and evaluation increases substantially as the number of agents grows. Considering the high computational cost when the number of robots scales up, we focus on two-agent settings in this work and leave coordination at larger team scales for future study.

\subsection{Bridging High-Level Reasoning and Low-Level Control via Natural Language Based Semantic Communication}
\label{sec:bridge}

Hierarchical robotic architectures bridge high-level reasoning and low-level control through different intermediate representations, including natural-language subgoals or motion descriptions~\citep{shi2025hi,belkhale2024rt}, spatial representations such as trajectories or visual subgoals~\citep{li2025hamster,zhao2025cot}, and learned latent representations in tightly coupled dual-system models~\citep{chen2025fast,bjorck2025gr00t}. Among these approaches, natural language offers interpretability and broad compatibility with language-conditioned models, whereas other representations can convey compact information but typically require stronger coupling between the two modules, making them harder to replace or adapt across architectures and embodiments.

Our framework adopts natural language as a unified interface between the high-level orchestrator and the low-level action model. The orchestrator harnesses the action model by generating low-level instructions in natural language, thereby decoupling high-level reasoning from low-level control. This design allows existing VLMs and VLAs to be incorporated as off-the-shelf components with minimal adaptation, without requiring them to be jointly trained through a specialized interface. It also enables either component to be independently replaced as stronger reasoning or action models become available.

Natural language further provides a common interface for inter-agent semantic communication. Each orchestrator can interpret messages from other agents even when they use different reasoning models, facilitating coordination across heterogeneous agents. The same interface also supports human participation by allowing users to inspect or revise instructions using ordinary language. More broadly, this shared linguistic interface provides a scalable foundation for future multi-robot systems, where heterogeneous robots and multiple human users can coordinate through a common communication protocol.

\subsection{Inter-Agent Information Sharing}
\label{sec:information sharing}

Distributed multi-agent systems face the challenges of partial observability and non-stationarity: each agent has only partial knowledge of the global task state, while the environment continues to evolve as other agents act. Insufficient awareness of such information can lead to conflicting actions or ineffective collaboration. To address these challenges, \ABBR{} enables agents to share and integrate structured natural-language messages during high-level reasoning. These messages convey key information about other agents' decisions and task-relevant changes in the environment, providing additional context for more precise high-level reasoning. To support effective decision-making, we structure each shared message using four fields:

\textbf{Intention.} It is identical to the low-level instruction sent to the action model and describes the next subtask that the ego agent intends to execute. Sharing this intention makes other agents aware of the ego agent's next action, helping them avoid behaviors that may interfere with its execution.

\textbf{Subgoals.} This field specifies the expected task state after the ego agent completes its current subtask. It explicitly communicates the future state that the ego agent aims to achieve, allowing other agents to act proactively to facilitate task completion or avoid undesirable outcomes.

\textbf{Belief of task.} This part describes task-relevant information observed or inferred by the ego agent, such as the location of a key object or whether a particular subtask has been completed. For example, in a kitchen task, a robot operating near the cabinets may inform another robot preparing to cook that ``the pot is in the third cabinet.'' Sharing such information reduces the impact of partial observability by propagating locally available knowledge across agents.

\textbf{Uncertainty of belief.} This optional field communicates the uncertainty associated with the ego agent's task belief. It is particularly useful in complex tasks, where information inferred through reasoning may not always be reliable. This uncertainty provides an additional confidence signal that allows the orchestrator to appropriately weigh different beliefs during high-level reasoning, especially when agents hold inconsistent beliefs or when the available information is ambiguous.

Through this structured information sharing, \ABBR{} provides each agent with compact contextual information about the intentions, subgoals, beliefs, and associated uncertainty of other agents. This additional context enables the orchestrator to generate more precise low-level instructions, thereby improving the safety and efficiency of multi-robot coordination.

\section{\BENCH{}}
\label{sec:benchmark}


\BENCH{} is designed to facilitate research on multi-robot manipulation by providing challenging long-horizon tasks with explicit cooperation requirements. Unlike RoboTwin \cite{chen2025robotwin}, which primarily targets dual-arm manipulation with two arms on the same embodiment and does not necessarily require explicit cooperation, \BENCH{} considers two independent robots and tasks specifically designed around multi-robot coordination. These tasks are either infeasible or inefficient for a single robot to complete independently, requiring each robot to reason and determine when cooperation is needed, and act at the appropriate time.
 
\BENCH{} highlights distributed execution, where each robot acts based on its own observations and policy while cooperating with another robot. The benchmark is built upon the ManiSkill3 framework and contains seven multi-robot manipulation tasks. In this section, we describe the task suite, basic settings, and dataset generation.

\subsection{Tasks for Multi-Robot Coordination}

All tasks share a common workspace configuration with two robot arms positioned on opposite sides of a table. The workspace and object placement are designed such that a single robot cannot always complete the task independently or efficiently, making cooperation necessary. \BENCH{} contains the following seven tasks: Food Serve, Prepare Snack, Clean Table, Cook Pot, Put Object Cabinet, Hang Bag, Exchange Bread. We further detail the task description in \ref{sec:task}.

\subsection{Observations, Actions, and Robots}

Each environment contains three RGB cameras: one global camera with a resolution of $512 \times 384$ and two wrist cameras with a resolution of $128 \times 128$. The global camera is positioned beside the table to provide an overview of the workspace, while each robot is equipped with a wrist camera that provides an egocentric view of its gripper. We adopt a distributed observation setting in which each robot receives observations from the global camera and its own wrist camera, but has no access to the wrist camera of the other robot. This design follows the distributed system setting and requires the robots to coordinate using only their locally available observations.

Both robots use 7-DoF Franka Emika Panda manipulators equipped with grippers. Under the distributed control setting, each policy controls only one robot rather than directly generating the joint actions of both robots. To improve task diversity, key objects are randomized within predefined task-specific regions at the beginning of each episode.

\subsection{Datasets}

We construct the datasets using the motion-planning-based trajectory generation pipeline provided by ManiSkill3. For each task, we first decompose the full manipulation procedure into a sequence of key states, such as grasping, releasing, or moving toward a target location. Given these key states, the generation script invokes the motion planner to construct smooth robot trajectories between consecutive states. The spatial and temporal constraints for each trajectories are carefully designed with the planner to generate the desired multi-robot coordination demonstrations.

In addition to robot trajectories, we associate each manipulation stage with a low-level instruction for each robot describing the current behavior. We also provide a high-level instruction that specifies the overall task objective. During data generation, the pipeline first plans a successful trajectory and records the corresponding robot actions and states. It then replays the trajectory in simulator to render the RGB camera observations. The resulting dataset therefore contains visual observations, states, robot actions, high-level, and low-level language instructions for both robots.  We further open-source the dataset to foster research on multi-robot coordination.

\section{Experiments}
\label{sec:experiment}

We evaluate \ABBR{} on 7 tasks from our \BENCH{} and 8 dual-arm tasks from RoboTwin 2.0 \citep{chen2025robotwin}, both under distributed control. Since RoboTwin originally adopts centralized control over the dual-arm robot, we adapt its setting by treating the two arms as independent robots and control them separately. Task overview are shown in \cref{fig:maniskill_task} and \cref{fig:robotwin_task}. All the finetuning and evaluation tasks are conducted on 4 NVIDIA A100 GPU, each taking 10--24 hours.  

In the \BENCH{} experiments, each Franka Emika Panda robot has 7 DoF and is equipped with a wrist camera, while both robots share a global camera positioned beside the workspace. We adopt the same distributed setting in which each robot has its own observations and policy. For each task, we collect 50 dual-robot demonstrations using planning-based trajectory generation and then decompose them into a unified single-robot dataset containing 100 trajectories for finetuning.

On RoboTwin, we select the tasks that require the coordination of two arms for multi-robot experiment. We treat the dual-arm Aloha-AgileX as two individual robots, each operating with its own observations and independently controlled by a separate policy. Each arm has 6 DoF with a wrist camera and a shared head camera. At finetuning stage, we use 50 official dual-arm demonstrations on each task. Each demonstration is decomposed into separate left- and right-arm trajectories, where each trajectory is paired with observations from the corresponding wrist camera and the shared head camera. We then merge the trajectories from both arms into a unified single-arm dataset, resulting in 100 demonstrations per task.

\subsection{Experiment Settings}

\textbf{Low-Level Control.} We adopt $\pi_{0.5}$ \citep{black2025pi05} as the low-level VLA controller in \ABBR{} due to its strong grounding in single-arm manipulation and generalization capability across diverse tasks and environments. We fine-tune $\pi_{0.5}$ using LoRA with batch sizes ranging from 256 to 1024 for 2k--10k training steps on \BENCH{} and with batch sizes 32 for 20k training steps on RoboTwin, depending on the task. Detailed fine-tuning configurations are provided in \ref{sec:config}. 

In \BENCH{} experiments, the dataset includes low-level instructions, which are directly used to fine-tune the VLA. In RoboTwin experiments, however, the official dataset does not include low-level instructions and instead provides a high-level instruction, which we use during fine-tuning. At evaluation time, we retain the same hierarchical setting as in \BENCH{} and provide decomposed low-level instructions to the VLA. Although these instructions does not fully match the fine-tuning dataset, the decomposition still benefits execution through the inherent generalization capability of the pretrained VLA. We further elaborate on this effect in \cref{sec:performance}.

\textbf{High-Level Reasoning.} We integrate Qwen3-VL-4B-Instruct \citep{bai2025qwen3} as the high-level reasoning module in \ABBR{}, as it provides strong visual grounding while remaining compact enough for practical inference. To prevent the model from generating low-level instructions that exceed the instruction-following capability of the VLA policy or fall outside its training distribution, we restrict each generated instruction to a predefined set of admissible examples. This constraint improves the reliability and consistency of low-level execution. In addition, to prevent the system from remaining indefinitely at an intermediate stage, we introduce a simple progress mechanism that forces the high-level instruction to advance when reasoning stagnates.

\textbf{Baseline.} To compare against other control models, we fine-tune $\pi_{0.5}$ and $\pi_{0}$ \citep{black2024pi_0} on each task using the corresponding high-level instruction as baseline methods. Except for the instruction setting, all fine-tuning configurations are kept identical to those used for \ABBR{}.

\textbf{Evaluation.} During evaluation, \ABBR{} and all baseline methods receive the same high-level instruction and observations as input from the simulation environment. We evaluate \ABBR{} and the baselines over 400 rollouts per task and report the corresponding success rates in \cref{tab:robotwin_res} and \cref{tab:robomulti_res}. Detailed evaluation settings are provided in \ref{sec:config}. 

{
\setlength{\floatsep}{1pt}      
\setlength{\textfloatsep}{4pt}  

\begin{table}[t]
\centering
\small
\caption{Comparison of success rates on \BENCH{} tasks.}
\label{tab:robomulti_res}

\begin{tabular*}{0.9\linewidth}{@{\extracolsep{\fill}}l|cccc}
\toprule
\textbf{Method}
& \textbf{Hang Bag}
& \textbf{Food Serve}
& \textbf{Prepare Snack}
& \textbf{Clean Table} \\
\midrule
\textbf{\ABBR{} w/ $\boldsymbol{\pi}_{0.5}$}
& \textbf{78.00}
& \textbf{32.75}
& \textbf{18.00}
& \textbf{24.75} \\

$\boldsymbol{\pi}_{0.5}$ only
& 52.25
& 29.75
& 16.25
& 16.50 \\

$\boldsymbol{\pi}_{0}$ only
& 58.75
& 22.25
& 3.75
& 0.00 \\
\bottomrule
\end{tabular*}

\vspace{0.5mm}

\begin{tabular*}{0.9\linewidth}{@{\extracolsep{\fill}}l|ccc}
\toprule
\textbf{Method}
& \textbf{Cook Pot}
& \textbf{Put Object Cabinet}
& \textbf{Exchange Bread} \\
\midrule
\textbf{\ABBR{} w/ $\boldsymbol{\pi}_{0.5}$}
& \textbf{39.25}
& \textbf{26.50}
& \textbf{31.00} \\

$\boldsymbol{\pi}_{0.5}$ only
& 1.00
& 18.50
& 17.50 \\

$\boldsymbol{\pi}_{0}$ only
& 0.25
& 14.50
& 18.50 \\
\bottomrule
\end{tabular*}

\end{table}

\begin{table}[t]
\centering
\small
\caption{Comparison of success rates on RoboTwin tasks.}
\label{tab:robotwin_res}

\begin{tabular*}{0.97\linewidth}{@{\extracolsep{\fill}}l|cccc}
\toprule
\textbf{Method}
& \textbf{Hanging Mug}
& \textbf{Pick Diverse Bottles}
& \textbf{Stack Bowls Two}
& \textbf{Put Bottle Dustbin}\\
\midrule
\textbf{\ABBR{} w/ $\boldsymbol{\pi}_{0.5}$}
& \textbf{24.75}
& \textbf{58.50}
& \textbf{85.50}
& \textbf{34.50}\\

$\boldsymbol{\pi}_{0.5}$ only
& 17.25
& 48.50
& 82.50
& 31.25\\

$\boldsymbol{\pi}_{0}$ only
& 12.50
& 42.50
& 33.75
& 21.25\\
\bottomrule
\end{tabular*}

\vspace{0.5mm}

\begin{tabular*}{0.97\linewidth}{@{\extracolsep{\fill}}l|cccc}
\toprule
\textbf{Method}
& \textbf{Lift Pot}
& \textbf{Handover Mic}
& \textbf{Blocks Ranking RGB}
& \textbf{Put Object Cabinet}\\
\midrule
\textbf{\ABBR{} w/ $\boldsymbol{\pi}_{0.5}$}
& \textbf{94.75}
& \textbf{98.00}
& \textbf{2.00}
& 43.25\\

$\boldsymbol{\pi}_{0.5}$ only
& 94.00
& 94.75
& 0.00
& \textbf{46.50}\\

$\boldsymbol{\pi}_{0}$ only
& 77.50
& 84.50
& 0.25
& 18.00\\
\bottomrule
\end{tabular*}

\end{table}
}

\subsection{Main Results}
\label{sec:performance}

We evaluate \ABBR{} on seven tasks in \BENCH{} and eight tasks in RoboTwin against strong baselines, with success rates reported in \cref{tab:robomulti_res} and \cref{tab:robotwin_res}, respectively. Detailed task rollout examples are provided in \cref{sec:rollout_example}. Overall, \ABBR{} matches or outperforms the baseline methods across most tasks. We attribute these gains to the high-level reasoning and inter-agent communication enabled by the hierarchical design.

For long-horizon multi-robot tasks, a global high-level instruction provides only coarse guidance throughout the entire task execution and offers limited information about what each robot should do at the current stage or how it should adapt to the behavior of its peers. With the hierarchical design, in contrast, the orchestrator reasons over the current task state together with messages from other agents, decomposes the task into appropriate subtasks, and provides specific low-level instructions to the action model. This allows each robot to adapt its behavior as the task evolves while avoiding conflicts with other agents.

\begin{wrapfigure}{l}{0.50\textwidth}
    \centering
    \includegraphics[width=\linewidth]{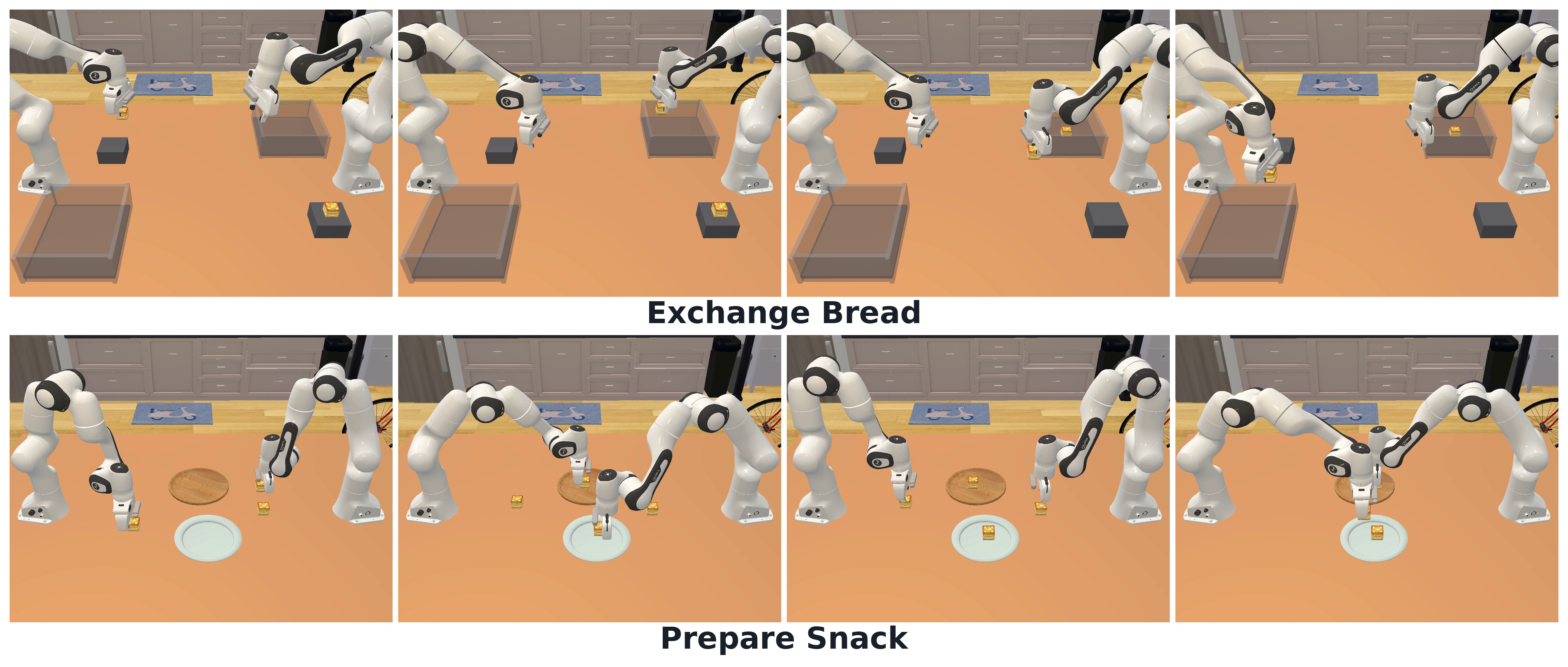}
    \vspace{-0.3in}
    \caption{We observe that explicit inter-agent communication enables the robots to better coordinate their behaviors, such as exchanging bread sequentially or placing objects on different plates. By avoiding conflicting or redundant actions in a narrow shared workspace, the robots achieve more efficient and safer coordination.}
    \label{fig:coordination}
    \vspace{-0.05in}
\end{wrapfigure}

Inter-agent communication is particularly important under partial observability, where local observations may not reveal the intentions or future actions of other robots. Explicit message sharing provides complementary information for high-level reasoning, enabling the orchestrator to make better-informed coordination decisions. With more precise low-level instructions, the action model can focus on executing the assigned subtask rather than implicitly reasoning about the overall task progression and multi-agent coordination. As shown in \cref{fig:coordination}, the robots coordinate effectively and complete the task in the proper sequence without conflicts.

This distinction becomes especially important when execution deviates from the planned behavior. Policies such as $\pi_0$ and $\pi_{0.5}$ that continuously follow a fixed high-level instruction may become uncertain about the appropriate next action after unexpected interactions or execution errors. Such situations occur frequently in multi-robot manipulation, where the behavior of one robot can alter another robot's workspace. The orchestrator can recognize these changes and adaptively issue new instructions, allowing the action model to recover and continue the task.

Additionally, tasks in \BENCH{} generally have longer horizons and require more stable coordination than those in RoboTwin, resulting in lower overall success rates. Nevertheless, the performance improvement of \ABBR{} over the baselines is more significant on \BENCH{}, further demonstrating the effectiveness of high-level reasoning and inter-agent communication and suggesting that their benefits become increasingly significant as the task horizon and coordination complexity increase.

\subsection{Ablation of Inter-Agent Communication}

\begin{figure*}[h]
    \centering

    \begin{subfigure}[t]{0.49\linewidth}
        \centering
        \includegraphics[height=1.75in,keepaspectratio]
        {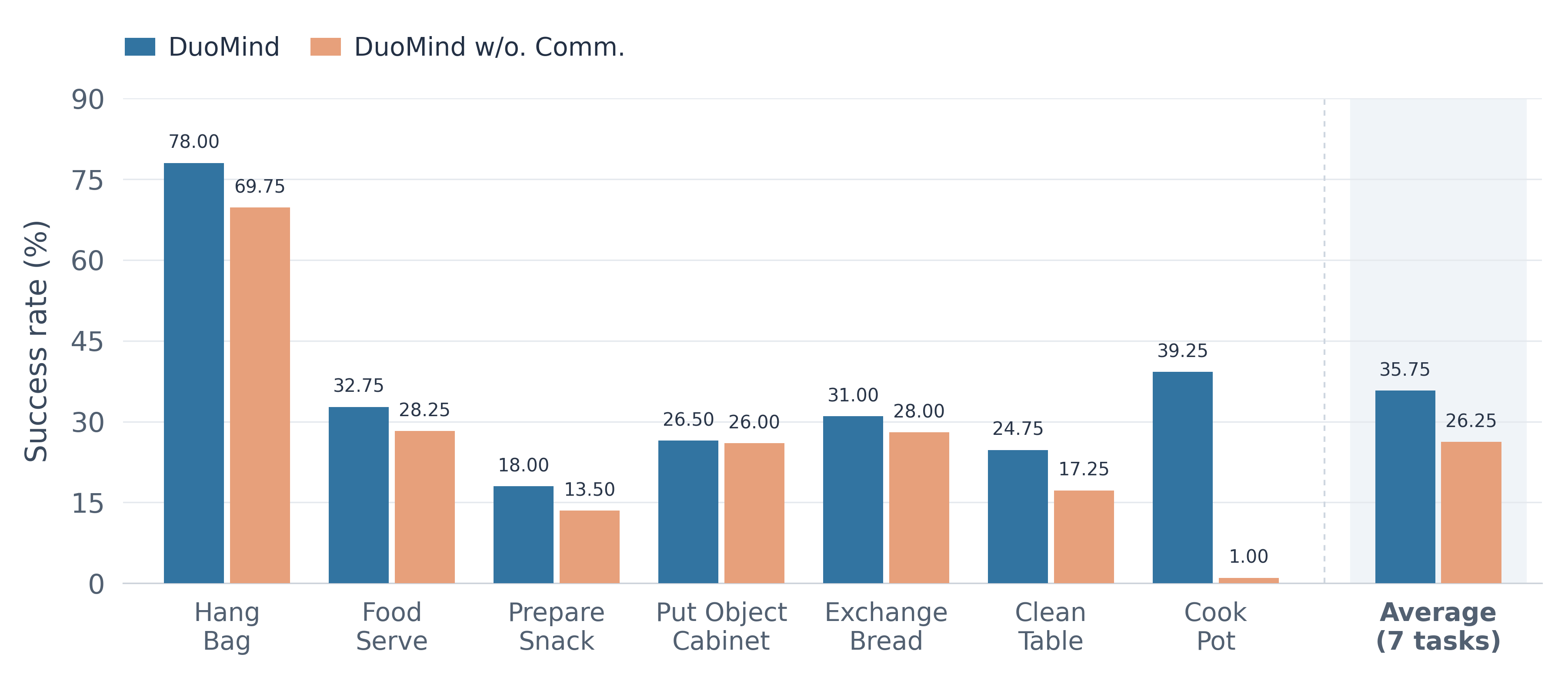}
        \vspace{-0.2in}
        \caption{Inter-agent communication ablation.}
        \label{fig:ab_comm}
    \end{subfigure}
    \hfill
    \begin{subfigure}[t]{0.49\linewidth}
        \centering
        \includegraphics[height=1.75in,keepaspectratio]
        {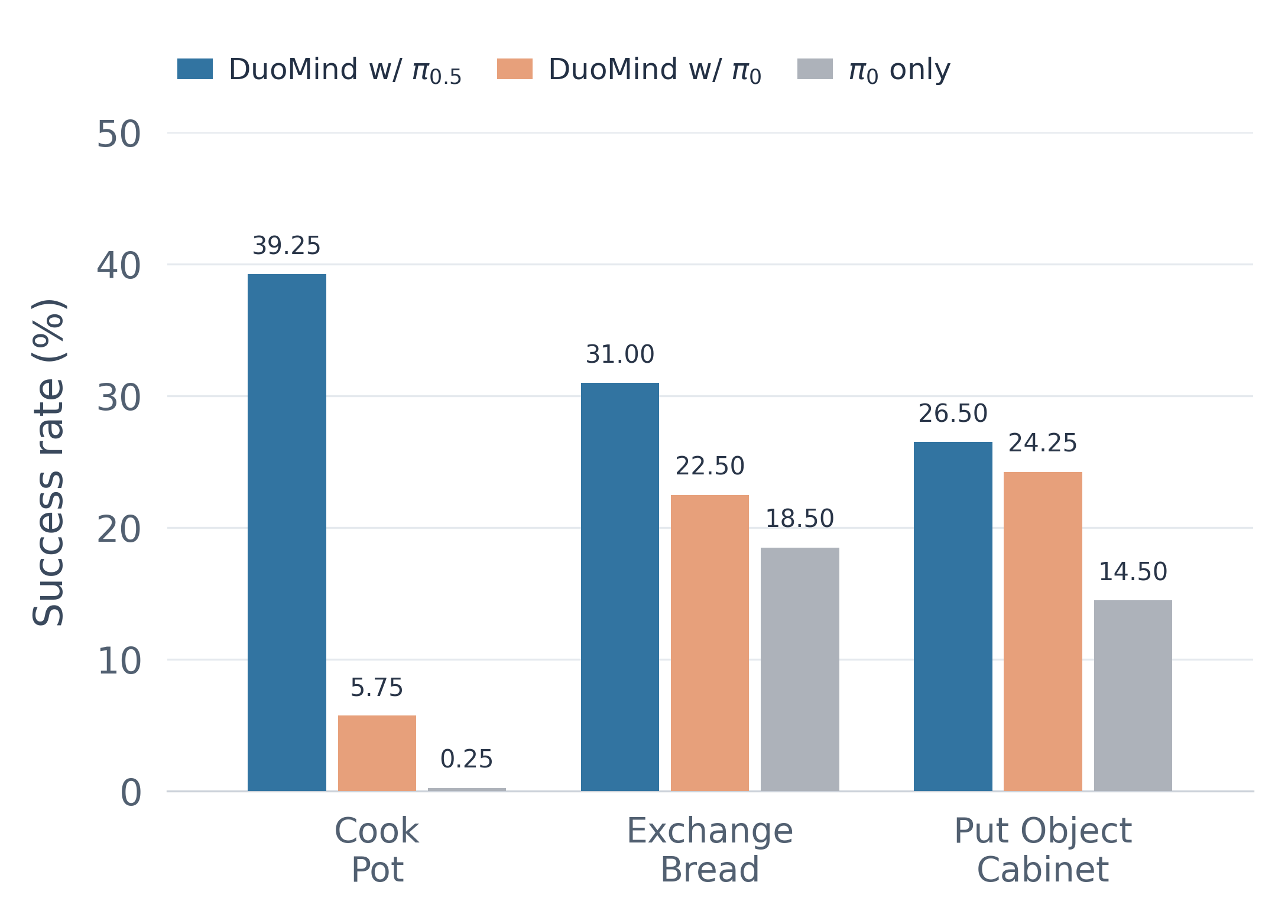}
        \vspace{-0.05in}
        \caption{Action model ablation.}
        \label{fig:ab_action_model}
    \end{subfigure}

    \vspace{-0.05in}
    \caption{
        Ablation studies of \ABBR{}.
        (a) Success rates when inter-agent communication is disabled.
        (b) Success rates with different action models.
    }
    \label{fig:ablation}
\end{figure*}

Inter-agent communication provides complementary information for decision making and helps mitigate partial observability. To evaluate its effect, we conduct an ablation, following the evaluation setting in \cref{sec:performance} except that messages from peer agents are disabled. As shown in \cref{fig:ab_comm}, removing inter-agent communication leads to more frequent conflicts and asynchronous behaviors, resulting in degraded coordination performance. These results demonstrate that explicitly sharing peer information is important for effective multi-robot coordination under distributed observations. We further elaborate the ablation study in \ref{sec:ap_comm}.

\subsection{Action Model Compatibility}
\ABBR{} decouples high-level orchestration from low-level control, allowing different action models to be integrated into the framework. To study the influence of the action model, we evaluate \ABBR{} with $\pi_{0}$ in place of $\pi_{0.5}$. As shown in \cref{fig:ab_action_model}, since $\pi_{0}$ is a weaker action model than $\pi_{0.5}$, \ABBR{} with $\pi_{0}$ achieves lower success rates than \ABBR{} with $\pi_{0.5}$, as expected. Nevertheless, it substantially outperforms the $\pi_{0}$-only baseline, demonstrating that the hierarchical orchestration of \ABBR{} consistently improves multi-robot coordination across different action models. At the same time, the stronger action model provides more reliable low-level execution, which further improves the overall task success rate. We detail this analysis in \ref{sec:ap_action_model}.

\section{Related Work}
\label{sec:related}

\subsection{Learning-Based Multi-Agent Coordination}

Classical multi-agent learning studies decentralized decision-making under partial observability and non-stationarity through centralized training, value decomposition, and learned communication
\citep{lowe2017multi,rashid2020monotonic,yu2022surprising,foerster2016learning}. These methods are effective in cooperative games and simulation, but often require task-specific joint training and encode coordination in latent representations that are hard to interpret or integrate with pretrained robot foundation models. Recent work extends coordination to physical robots through decentralized imitation learning, shared representations, and VLA policies
\citep{dong2025mimic,he2025latent,qin2025robofactory,doshi2026chorus}. In contrast, \ABBR{} uses decentralized reasoning with explicit natural-language communication for interpretable exchange of task-relevant information.

\subsection{Robot Learning and Multi-Robot Planning}

Generalist robot policies leverage large-scale robot data and pretrained vision/language representations to map observations or language instructions to actions
\citep{zitkovich2023rt,kim2024openvla,black2024pi_0, zhou2026agile, team2025gemini, gao2026vita}. Hierarchical approaches further separate high-level semantic reasoning from low-level continuous control
\citep{shi2025hi,black2025pi05,song2025hume,team2025gemini}, but mainly focus on single-robot execution. LLM- and VLM-based multi-robot systems explore task decomposition, allocation, communication, feedback, and replanning
\citep{kannan2024smart,chen2024scalable,mandi2024roco,liu2025coherent,chen2025emos,wang2024dart,he2026closed}, while typically relying on predefined skills, symbolic actions, waypoints, or motion planners. \ABBR{} instead combines distributed high-level reasoning and communication with VLA-based physical execution.

\subsection{Robot Learning and Multi-Agent Benchmarks}

Established robot-learning benchmarks provide diverse manipulation tasks, demonstrations, language instructions, and long-horizon evaluation, but largely focus on single robots or centrally controlled embodiments
\citep{yu2020meta,james2020rlbench,gu2023maniskill2,mees2022calvin,liu2023libero}. Recent multi-agent benchmarks study collaborative planning, heterogeneous capabilities, and manipulation
\citep{mandi2024roco,liu2025coherent,chen2025emos,chang2025partnr,qin2025robofactory}. However, they often emphasize either abstract high-level planning or low-level manipulation. \BENCH{} instead evaluates long-horizon coordination and closed-loop execution under distributed observation and control.

\section{Conclusion}

We presented \ABBR{}, a distributed hierarchical framework for multi-robot coordination that combines high-level reasoning, low-level control, and semantic inter-agent communication. We further introduced \BENCH{}, a benchmark for evaluating long-horizon multi-robot manipulation under distributed observations and control. Experiments on \BENCH{} and RoboTwin demonstrate that hierarchical orchestration improves multi-robot task execution, while ablation studies highlight the importance of explicit information sharing for coordination under partial observability. The decoupled design also allows different reasoning and action models to be integrated without redesigning the overall framework. As stronger VLMs and VLAs continue to emerge, \ABBR{} can naturally incorporate these advances to further improve its coordination capabilities. Leveraging such advances toward more capable and efficient real-world multi-robot coordination is a promising direction for future work.

\subsubsection*{Acknowledgments}
We thank Keyu Zhu and Vivian Xie for their assistance in developing \BENCH{}.

\bibliography{iclr2027_conference}
\bibliographystyle{iclr2027_conference}

\appendix

\section{Task Settings of \BENCH{}}
\label{sec:task}

\BENCH{} contains the following seven tasks:

\textbf{Food Serve.}
Two pieces of bread and a serving tray are placed across the workspace such that one robot cannot reach both the bread and the tray. The robots must transfer the bread between each other when necessary and place both pieces onto the tray.

\textbf{Prepare Snack.}
Two plates are placed near the center of the table, while four pieces of bread are distributed across the two sides of the workspace. The robots must place two pieces on each plate. Both robots can operate simultaneously and must coordinate their motions to efficiently complete the task while avoiding collisions.

\textbf{Clean Table.}
Two cubes and a container are placed on the table such that neither robot can independently reach all cubes and container. Completing the task requires a cube to be passed between the robots when the container lies outside the workspace of a robot.

\textbf{Cook Pot.}
A carrot, a pot with a lid, and an oven are placed around the center of the workspace. The robots must first remove the lid, place the carrot into the pot, and replace the lid. They should then jointly grasp the two handles of the pot and transport it to the oven.

\textbf{Put Object Cabinet.}
A cube and a cabinet are placed in the workspace. The robots must open the bottom drawer of the cabinet, place the cube inside, and close the drawer. The task requires coordination between two robot arms for higher efficiency.

\textbf{Hang Bag.}
A bag and a hook are placed on opposite sides of the workspace such that neither robot can reach both. One robot must first pass the bag to the other robot, which then hangs the bag onto the hook.

\textbf{Exchange Bread.}
A piece of bread and a target box are placed beside each robot arm. Each robot must place its own bread into the box on the opposite side of the table. Since the target box lies outside its reachable workspace, the robot must first hand the bread over to the other robot, which then places it into the corresponding box.

\begin{figure*}[h]
    \centering
    \includegraphics[width=1.0\linewidth]{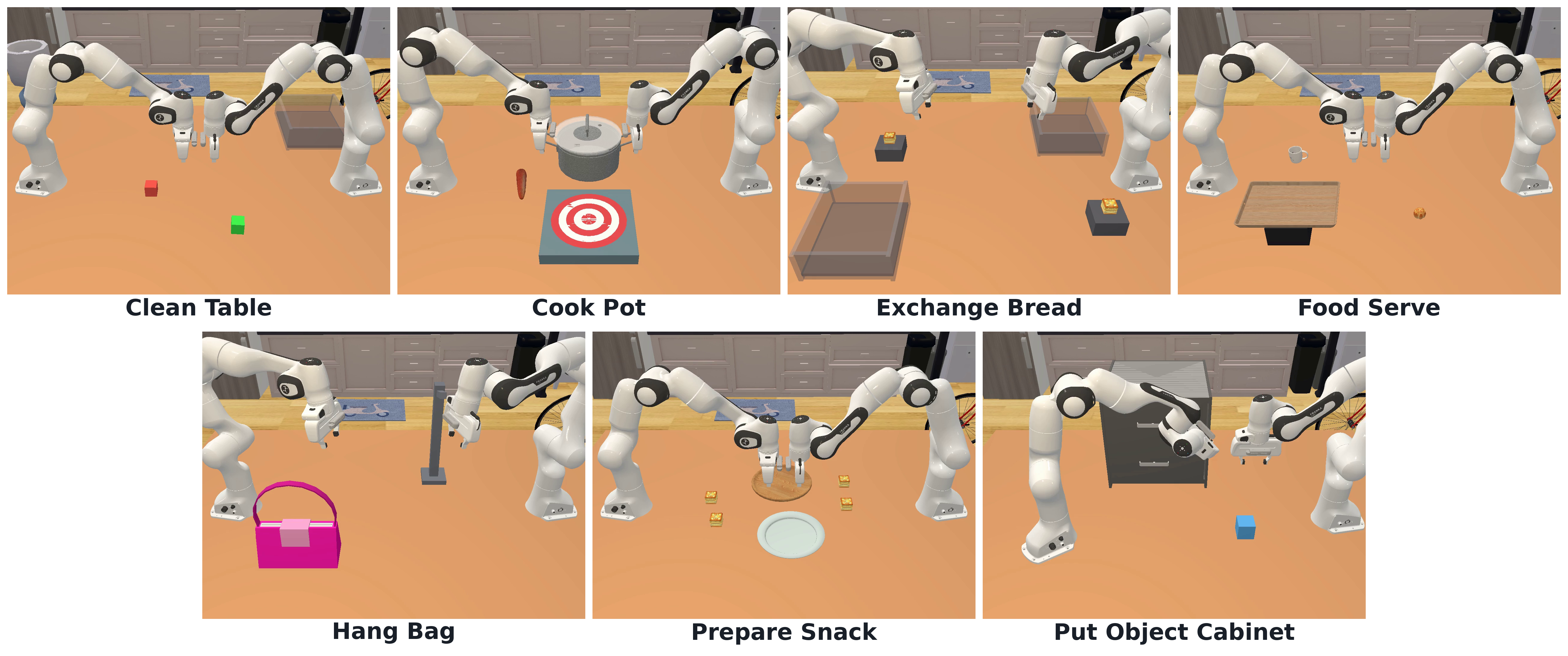}
    \vspace{-0.3in}
    \caption{The task overview of \BENCH{}.}
    \label{fig:maniskill_task}
\end{figure*}

\begin{figure*}[h]
    \centering
    \includegraphics[width=1.0\linewidth]{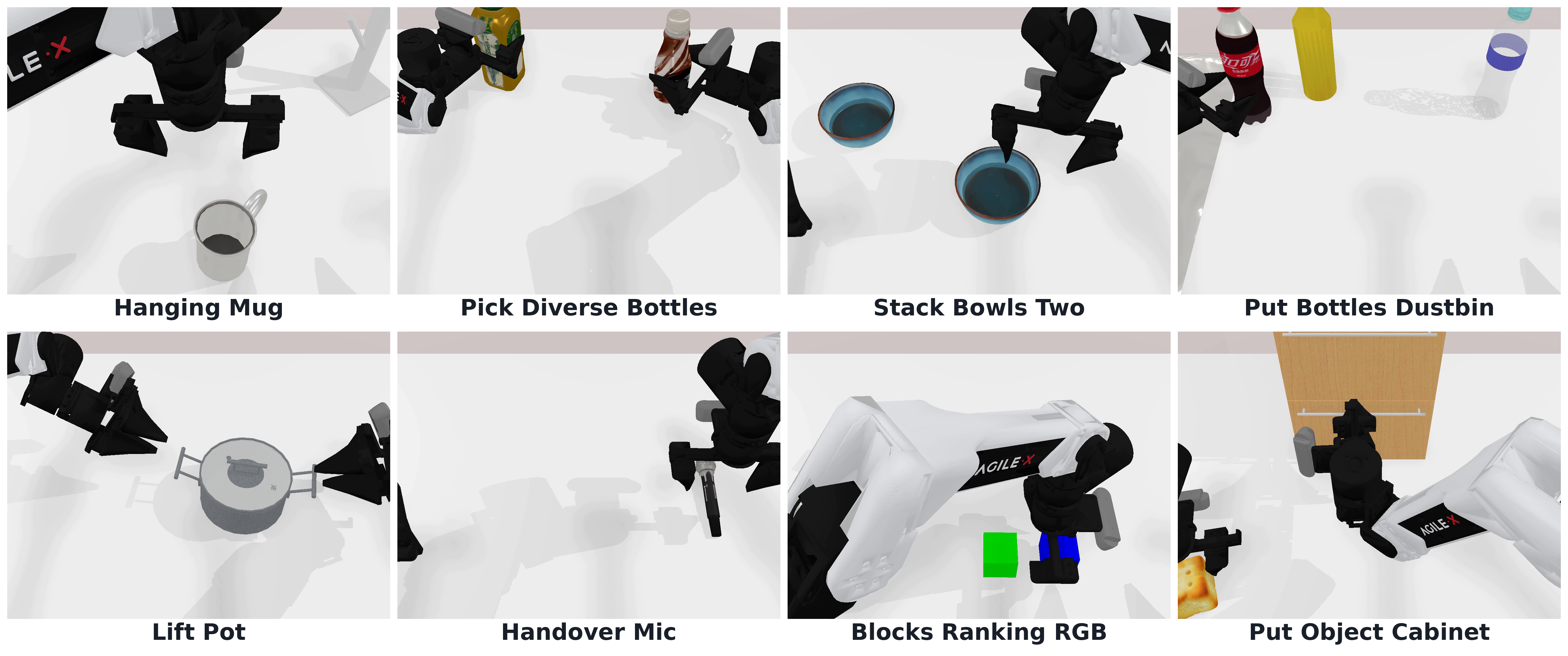}
    \vspace{-0.3in}
    \caption{The task overview of selected RoboTwin tasks.}
    \label{fig:robotwin_task}
\end{figure*}

\section{Effect of Inter-Agent Communication}
\label{sec:ap_comm}

\begin{figure*}[h]
    \centering
    \includegraphics[width=1.0\linewidth]{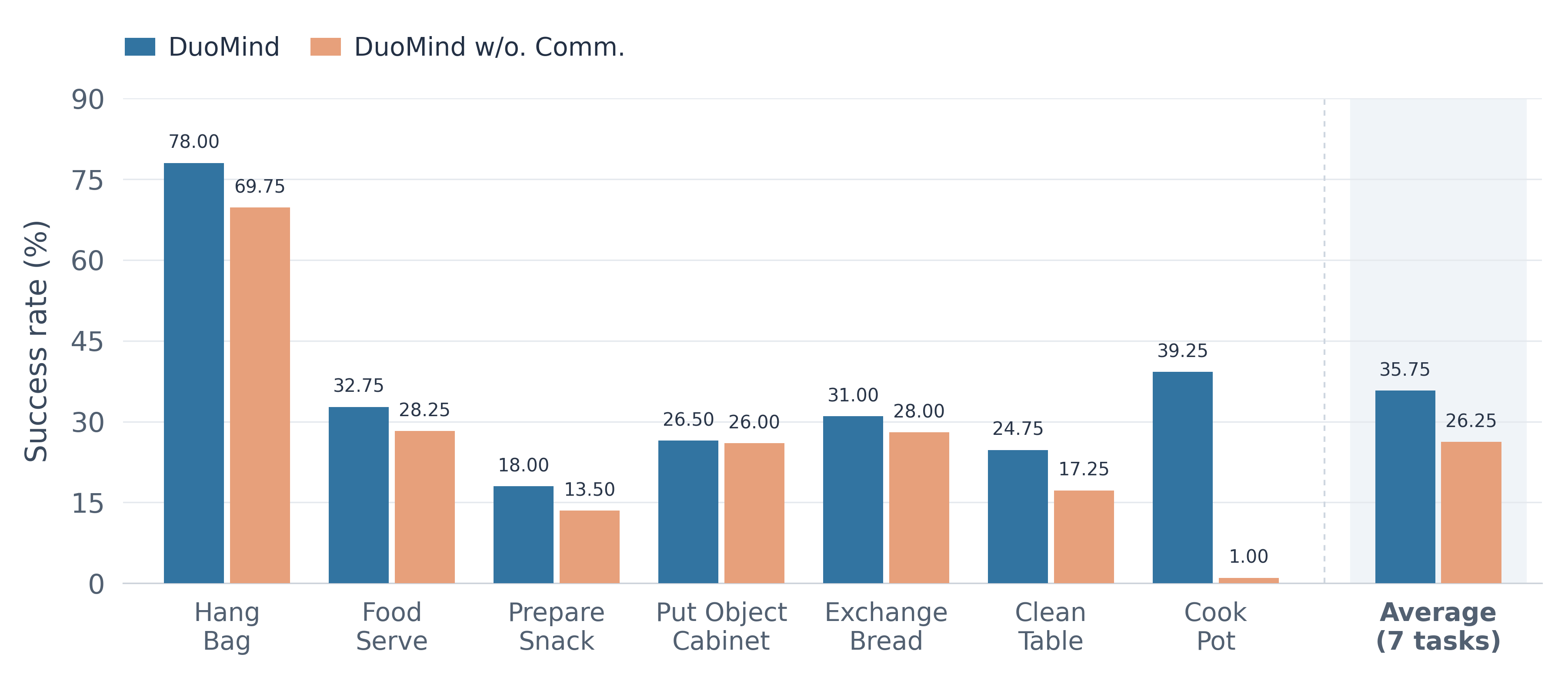}
    \vspace{-0.3in}
    \caption{The result of ablation study that disable the inter-agent communication in \ABBR{}.}
    \label{fig:ap_comm}
\end{figure*}

Inter-agent communication shares complementary information for decision making and helps mitigate the partial observability inherent in distributed multi-robot settings. To further study the effect of the shared messages, we conduct an ablation using \ABBR{} with $\pi_{0.5}$ as the action model, following the evaluation setting in \cref{sec:performance} except that messages from peer agents are disabled. Under this setting, each orchestrator receives no information from the other agent during reasoning and generates low-level instructions based only on its local observation and the task instruction.

The results are reported in \cref{fig:ap_comm}. Without inter-agent communication, we observe more frequent conflicts and asynchronous behaviors between the two robots. For example, in \textit{Cook Pot}, the two robots are required to lift the pot handles in a coordinated manner. Without communication, one robot often starts lifting while the other has only begun grasping the handle, causing the coordinated manipulation to fail. In \textit{Prepare Snack}, the two robots may independently select the same plate as their target and attempt to place bread simultaneously, which can lead to collisions between the robot arms above the plate.

These failure cases illustrate a limitation of relying solely on local observations: an individual agent may not have sufficient information to infer the peer agent's current intention or upcoming behavior. Explicit inter-agent communication provides a direct and efficient mechanism for sharing such information, thereby reducing conflicts and improving synchronized and safe multi-robot coordination.

\Needspace{10\baselineskip}
\section{Action Model Compatibility}
\label{sec:ap_action_model}

\Needspace{0.45\textheight}
\begin{wrapfigure}{l}{0.60\textwidth}
    \centering
    \includegraphics[width=\linewidth]{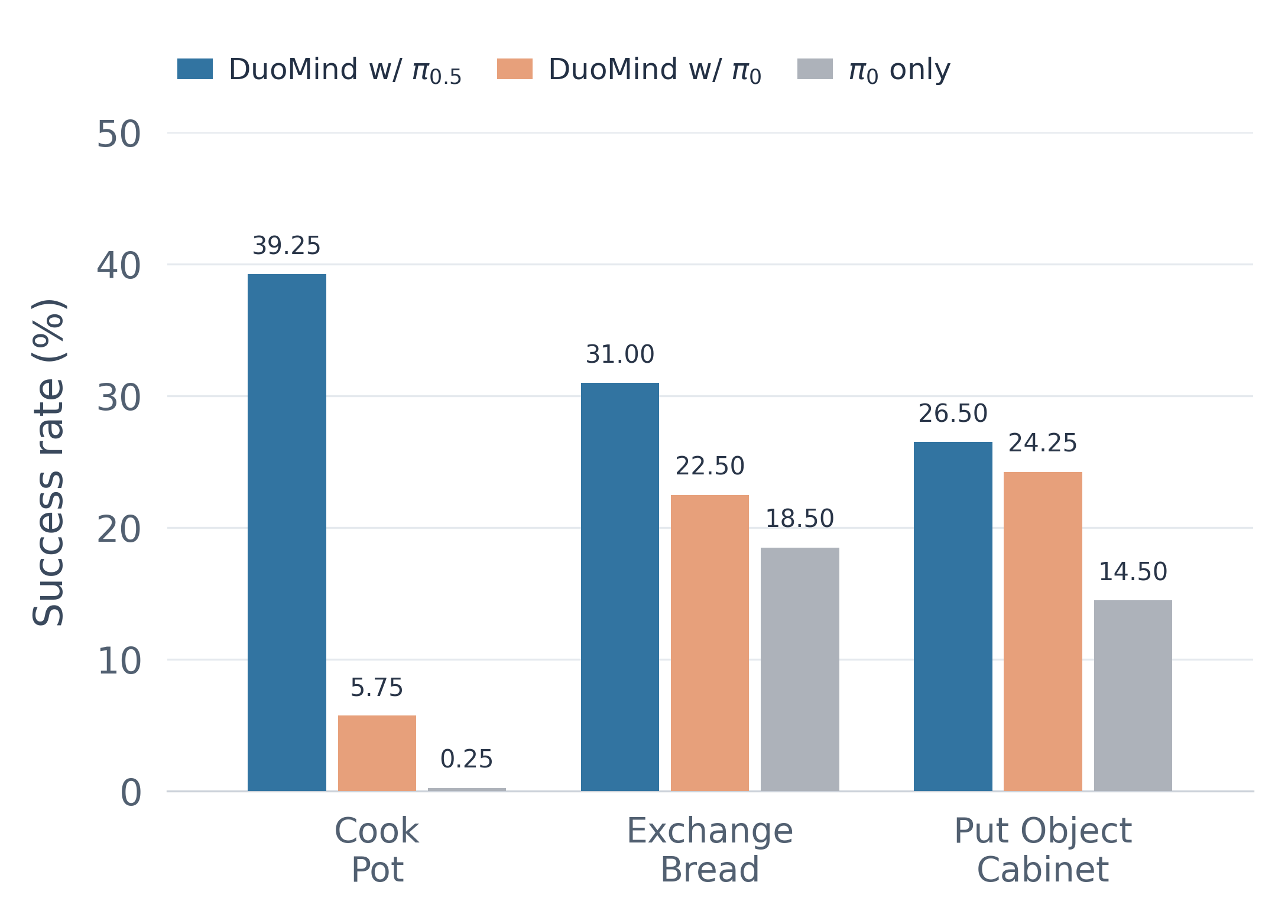}
    \vspace{-0.3in}
    \caption{Results of the ablation study using different action models in \ABBR{}.}
    \label{fig:ap_action_model}
\end{wrapfigure}

\ABBR{} decouples the high-level orchestrator from the low-level action model, enabling different action models to be integrated without changing the overall coordination framework. To analyze the influence of the action model, we conduct an additional experiment using $\pi_{0}$ in place of $\pi_{0.5}$, while following the same fine-tuning procedure described in \cref{sec:performance}.

As shown in \cref{fig:ap_action_model}, because $\pi_{0}$ provides weaker low-level control capability than $\pi_{0.5}$, \ABBR{} with $\pi_{0}$ achieves lower success rates than \ABBR{} with $\pi_{0.5}$, as expected. However, \ABBR{} with $\pi_{0}$ still substantially outperforms the $\pi_{0}$-only baseline. This comparison shows that the performance gain is not solely determined by the underlying action model. By introducing high-level reasoning and subtask decomposition through the orchestrator, \ABBR{} provides more appropriate low-level instructions for the evolving multi-agent state and improves coordination over the action-model-only setting.

The comparison also highlights the complementary roles of the two components. The orchestrator improves high-level reasoning and multi-agent coordination, while a stronger action model provides more reliable and stable low-level execution. Consequently, improving the capability of the action model can further increase the overall success rate. These results demonstrate that \ABBR{} is compatible with different action models while continuing to benefit from advances in low-level VLA policies.

\Needspace{9\baselineskip}
\section{Training and Evaluation Configurations}
\label{sec:config}

The fine-tuning and evaluation settings of \BENCH{} and RoboTwin are summarized in
\cref{tab:robopoly_config,tab:robotwin_config}. All methods use LoRA fine-tuning and 400 evaluation rollouts per experiment. The batch sizes reported in the table refer to global batch size. For RoboTwin, the policy takes the head camera and each robot's own wrist image as input, while on \BENCH{} it takes the global camera and each robot's own wrist image. The fine-tuning and
evaluation of $\pi_{0.5}$ and $\pi_0$ adopt the same settings.

In the tables, \textit{Execute / horizon} denotes the number of actions executed from each predicted action chunk and the total action-chunk horizon, respectively. \textit{VLM interval} denotes the number of policy chunks
executed between two consecutive calls to the high-level VLM orchestrator. We use a linear warmup over the first 1,000 training steps to reach the peak learning rate, followed by cosine decay to a final learning rate of $2.5\times10^{-6}$.

\begin{table}[H]
\centering
\caption{Fine-tuning and evaluation settings on \BENCH{}.}
\label{tab:robopoly_config}

\small
\setlength{\tabcolsep}{4.2pt}
\renewcommand{\arraystretch}{1.12}

\begin{tabular}{llccccc}
\toprule
\textbf{Task}
& \textbf{Method}
& \textbf{Train steps}
& \textbf{Batch}
& \textbf{Peak LR}
& \textbf{Execute / horizon}
& \textbf{VLM interval} \\
\midrule

Hang Bag
& $\pi_{0.5}$/$\pi_0$ only
& 10k & 256 & $2.5\times10^{-5}$ & 12/16 & -- \\
Hang Bag
& \ABBR{}
& 10k & 256 & $2.5\times10^{-5}$ & 12/16 & 4 \\

\addlinespace[2pt]

Food Serve
& $\pi_{0.5}$/$\pi_0$ only
& 10k & 256 & $2.5\times10^{-5}$ & 16/16 & -- \\
Food Serve
& \ABBR{}
& 10k & 256 & $2.5\times10^{-5}$ & 16/16 & 4 \\

\addlinespace[2pt]

Prepare Snack
& $\pi_{0.5}$/$\pi_0$ only
& 2k & 512 & $1\times10^{-4}$ & 16/16 & -- \\
Prepare Snack
& \ABBR{}
& 2k & 512 & $1\times10^{-4}$ & 16/16 & 4 \\

\addlinespace[2pt]

Put Object Cabinet
& $\pi_{0.5}$/$\pi_0$ only
& 4k & 512 & $1\times10^{-4}$ & 8/16 & -- \\
Put Object Cabinet
& \ABBR{}
& 4k & 512 & $1\times10^{-4}$ & 8/16 & 4 \\

\addlinespace[2pt]

Exchange Bread
& $\pi_{0.5}$/$\pi_0$ only
& 4k & 512 & $1\times10^{-4}$ & 16/16 & -- \\
Exchange Bread
& \ABBR{}
& 4k & 512 & $1\times10^{-4}$ & 16/16 & 3 \\

\addlinespace[2pt]

Clean Table
& $\pi_{0.5}$/$\pi_0$ only
& 4k & 512 & $1\times10^{-4}$ & 16/16 & -- \\
Clean Table
& \ABBR{}
& 4k & 512 & $1\times10^{-4}$ & 16/16 & 4 \\

\addlinespace[2pt]

Cook Pot
& $\pi_{0.5}$/$\pi_0$ only
& 4k & 512 & $1\times10^{-4}$ & 16/16 & -- \\
Cook Pot
& \ABBR{}
& 4k & 512 & $1\times10^{-4}$ & 16/16 & 4 \\

\bottomrule
\end{tabular}
\end{table}

\begin{table}[H]
\centering
\caption{Fine-tuning and evaluation settings on RoboTwin.}
\label{tab:robotwin_config}

\small
\setlength{\tabcolsep}{4.2pt}
\renewcommand{\arraystretch}{1.12}

\begin{tabular}{llccccc}
\toprule
\textbf{Task}
& \textbf{Method}
& \textbf{Train steps}
& \textbf{Batch}
& \textbf{Peak LR}
& \textbf{Execute / horizon}
& \textbf{VLM interval} \\
\midrule

Hanging Mug
& $\pi_{0.5}$/$\pi_0$ only
& 20k & 32 & $2.5\times10^{-5}$ & 50/50 & -- \\
Hanging Mug
& \ABBR{}
& 20k & 32 & $2.5\times10^{-5}$ & 50/50 & 4 \\

\addlinespace[2pt]

Pick Diverse Bottles
& $\pi_{0.5}$/$\pi_0$ only
& 20k & 32 & $2.5\times10^{-5}$ & 50/50 & -- \\
Pick Diverse Bottles
& \ABBR{}
& 20k & 32 & $2.5\times10^{-5}$ & 50/50 & 4 \\

\addlinespace[2pt]

Stack Bowls Two
& $\pi_{0.5}$/$\pi_0$ only
& 20k & 32 & $2.5\times10^{-5}$ & 50/50 & -- \\
Stack Bowls Two
& \ABBR{}
& 20k & 32 & $2.5\times10^{-5}$ & 50/50 & 2 \\

\addlinespace[2pt]

Put Bottles Dustbin
& $\pi_{0.5}$/$\pi_0$ only
& 20k & 32 & $2.5\times10^{-5}$ & 50/50 & -- \\
Put Bottles Dustbin
& \ABBR{}
& 20k & 32 & $2.5\times10^{-5}$ & 50/50 & 4 \\

\addlinespace[2pt]

Lift Pot
& $\pi_{0.5}$/$\pi_0$ only
& 20k & 32 & $2.5\times10^{-5}$ & 50/50 & -- \\
Lift Pot
& \ABBR{}
& 20k & 32 & $2.5\times10^{-5}$ & 50/50 & 4 \\

\addlinespace[2pt]

Handover Mic
& $\pi_{0.5}$/$\pi_0$ only
& 20k & 32 & $2.5\times10^{-5}$ & 50/50 & -- \\
Handover Mic
& \ABBR{}
& 20k & 32 & $2.5\times10^{-5}$ & 50/50 & 4 \\

\addlinespace[2pt]

Blocks Ranking RGB
& $\pi_{0.5}$/$\pi_0$ only
& 20k & 32 & $2.5\times10^{-5}$ & 50/50 & -- \\
Blocks Ranking RGB
& \ABBR{}
& 20k & 32 & $2.5\times10^{-5}$ & 50/50 & 4 \\

\addlinespace[2pt]

Put Object Cabinet
& $\pi_{0.5}$/$\pi_0$ only
& 20k & 32 & $2.5\times10^{-5}$ & 50/50 & -- \\
Put Object Cabinet
& \ABBR{}
& 20k & 32 & $2.5\times10^{-5}$ & 50/50 & 4 \\

\bottomrule
\end{tabular}
\end{table}

\section{Evaluation Visualization}
\label{sec:rollout_example}

\noindent

\begin{figure}[H]
    \centering
    \includegraphics[width=1.0\linewidth]{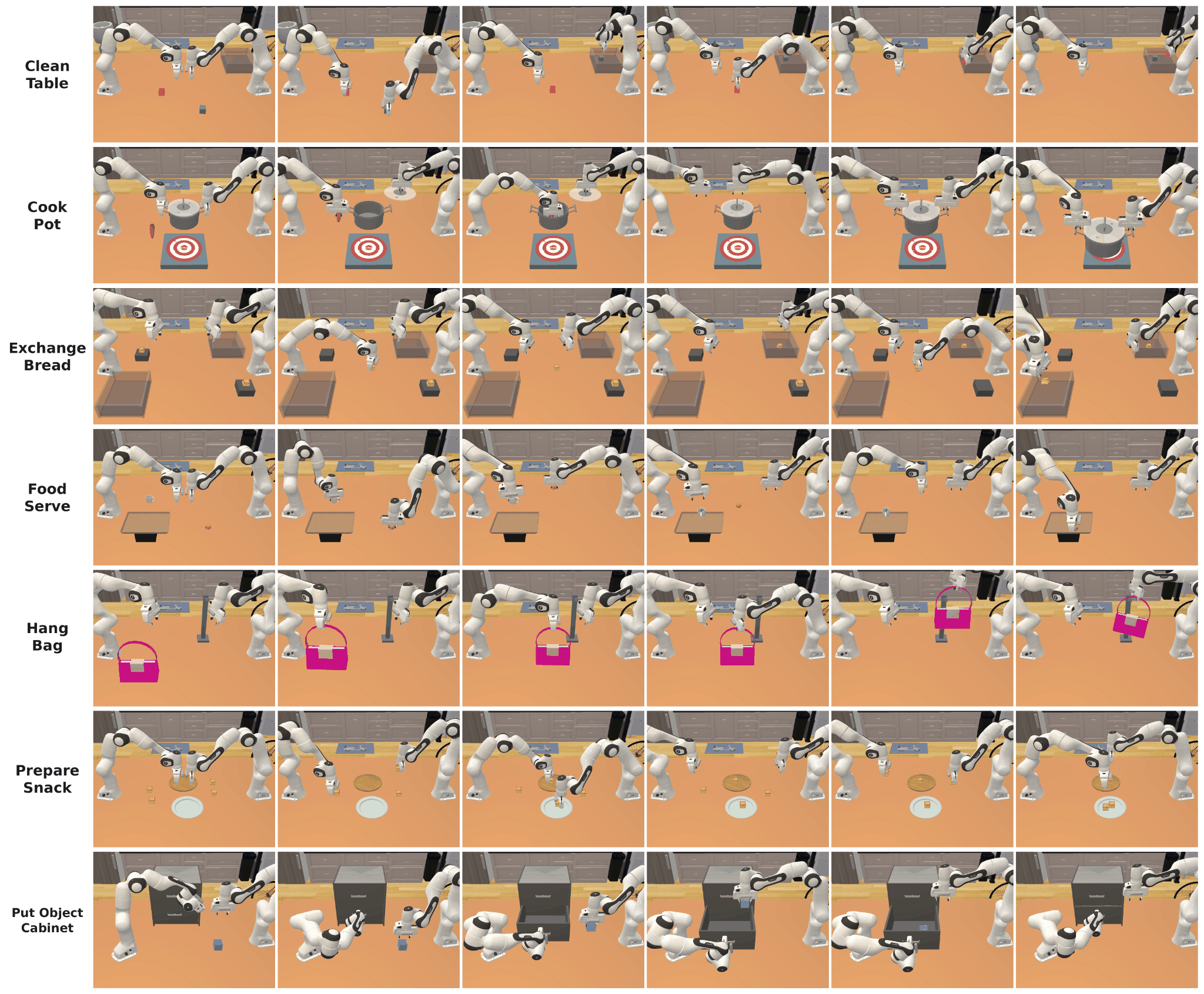}
    \vspace{-0.3in}
    \caption{We evaluate \ABBR{} on 7 \BENCH{} tasks.}
    \label{fig:all_eval}
\end{figure}

\begin{figure}[H]
    \centering
    \includegraphics[width=1.0\linewidth]{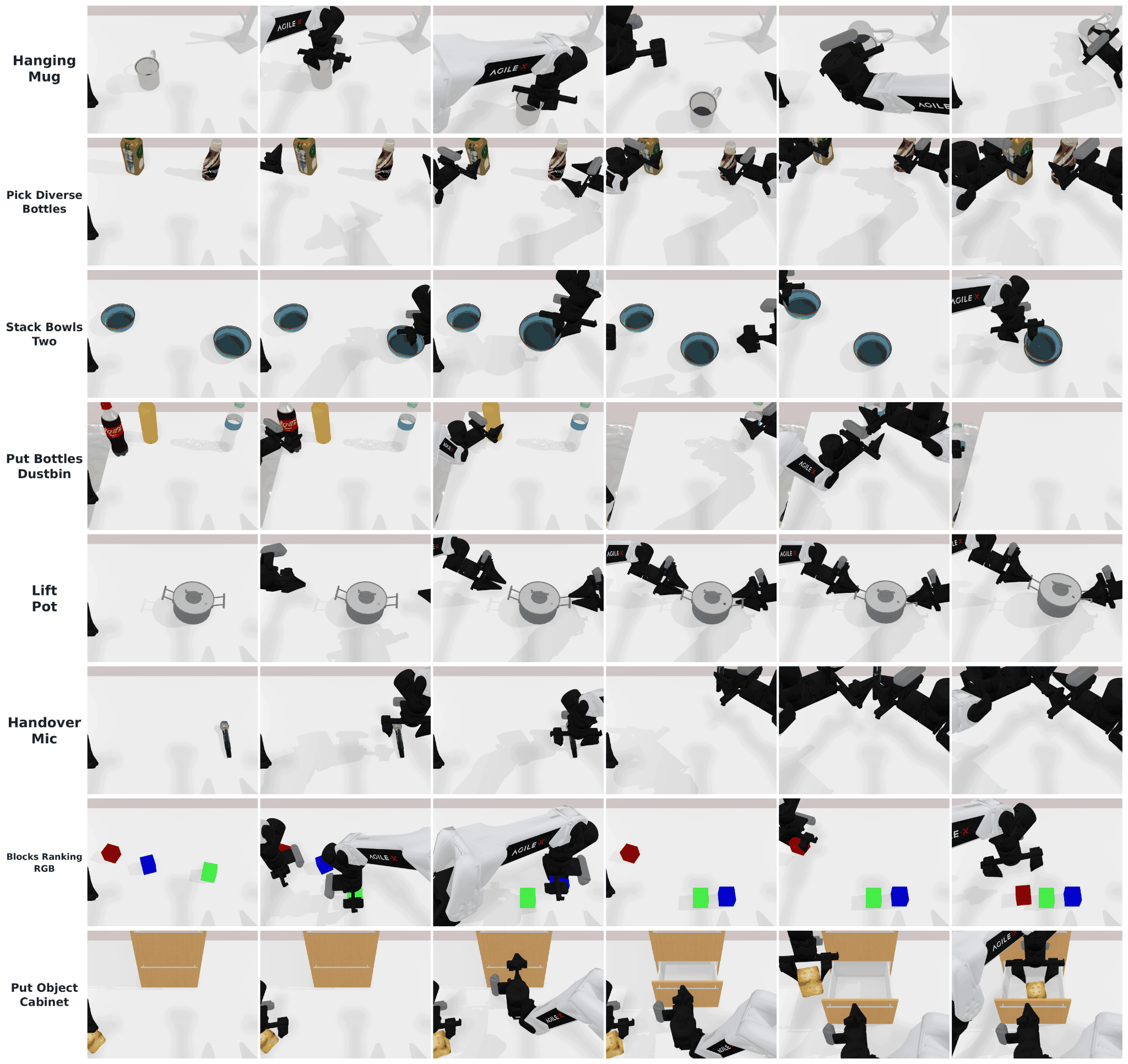}
    \vspace{-0.3in}
    \caption{We evaluate \ABBR{} on 8 selected RoboTwin tasks.}
    \label{fig:robotwin_all_eval}
\end{figure}

\section{VLM System Prompt}
\label{sec:prompt}

We guide the orchestrator with a general system prompt shared across agents and tasks, which specifies its role and the type of low-level instruction to generate for the action model. The orchestrator serves two roles. First, it performs high-level planning by reasoning over the ego agent’s reachable workspace, the peer agent’s intention, and the current task state. It selects a complementary subgoal that avoids conflicts with other agents and contributes effectively to the overall task. Second, it generates a structured message for inter-agent communication, providing concise information about the ego agent’s intention, belief, and uncertainty to support the peer agent’s reasoning and coordination.

\begin{tcolorbox}[
    title={System Prompt for Orchestrator},
    colback=gray!7,
    colframe=gray!90,
    fonttitle=\bfseries,
    boxsep=0.8mm,       
    left=1mm,
    right=1mm,
    top=1mm,
    bottom=1mm
]
\begin{Verbatim}[
    breaklines=true,
    breakanywhere=true,
    breaksymbolleft={},
    breaksymbolright={},
    fontsize=\scriptsize,
    baselinestretch=0.85,  % tighter line spacing
    frame=single,
    framesep=0.6mm         % text-to-inner-frame spacing
]
You are the system-2 planner for agent_<AGENT_ID>, controlling the <ARM_SIDE> arm (panda-<AGENT_ID>).
Use the global camera image, the high-level task, and the peer message to choose one short low-level instruction for this agent's VLA.

You receive the current global-camera image.
High-level instruction: <HIGH_LEVEL_INSTRUCTION>
This agent's currently accepted instruction: <PREVIOUS_ACCEPTED_INSTRUCTION>
This agent's own recent planning history (oldest to newest): <OWN_RECENT_HISTORY_JSON>
Other agent shared message: {"intention": "<PEER_INTENTION>", "subgoals": "<PEER_SUBGOALS>", "belief": "<PEER_BELIEF>"}

Without a specific requirement to act on a particular object or region, prefer objects and subgoals in this agent's reachable workspace and choose a target that complements the peer.

The low-level instruction must look like these training instructions:
<ALL_TRAINING_INSTRUCTION_EXAMPLES>

This ManiSkill task supplies an arm-specific candidate set. The L1/R1-style prefixes are metadata tags, not part of the low-level instruction. Select exactly one candidate, but put only the sentence after its colon in low_level_instruction; do not invent, paraphrase, include the tag, or select another arm's instruction.
Permitted candidates for this <ARM_SIDE> arm:
<THIS_ARM_CANDIDATES>

The numeric candidate suffixes are ordered stages for this arm. On the first planning event, select the lowest-numbered stage. On later events, repeat the current stage until its result is visibly complete, then advance by at most one stage. Never skip or return to an earlier stage.

Progress rule: do not keep issuing a subgoal that is already complete. Review the current image and your own history when available. If visual evidence or your own prior belief says the current subgoal's observable result is complete, you MUST select a different permitted candidate that advances the unfinished high-level task. Otherwise repeat the current candidate. Do not advance merely because time passed, and visually verify claims from the other agent.

Return only valid JSON with this exact schema:
{
  "low_level_instruction": "short command for this agent VLA",
  "shared_message": {
    "intention": "same as low_level_instruction",
    "subgoals": "observable condition when the low-level instruction is done",
    "belief": "task-relevant CURRENT observed state; never describe an intended future result as already complete"
  }
}
\end{Verbatim}
\end{tcolorbox}

\end{document}